\documentclass{article} 

\usepackage{PRIMEarxiv}
\usepackage[utf8]{inputenc} % allow utf-8 input
\usepackage[T1]{fontenc}    % use 8-bit T1 fonts
\usepackage{hyperref}       % hyperlinks
\usepackage{url}            % simple URL typesetting
\usepackage{booktabs}       % professional-quality tables
\usepackage{amsfonts}       % blackboard math symbols
\usepackage{nicefrac}       % compact symbols for 1/2, etc.
\usepackage{microtype}      % microtypography
\usepackage{fancyhdr}       % header
\usepackage{graphicx}       % graphics
\graphicspath{{media/}} 
\usepackage{amsmath}
\usepackage{amssymb}
\usepackage{amsthm}
\usepackage{natbib}
\usepackage{subfigure}

\title{Causal Discovery with Stage Variables for Health Time Series}

\author{
  Bharat Srikishan, Samantha Kleinberg \\
  Department of Computer Science\\
  Stevens Institute of Technology\\
  Hoboken, NJ\\
  \texttt{\{bsrikish, samantha.kleinberg\}@stevens.edu} \\
}

\usepackage{algorithm}
\usepackage{algorithmic}

\usepackage{tikz}
\usetikzlibrary{arrows, automata}

\usepackage{pifont}% http://ctan.org/pkg/pifont
\newcommand{\cmark}{\ding{51}}%
\newcommand{\xmark}{\ding{55}}

\newcommand{\bX}{\mathbf{X}}

\newcommand{\bS}{\mathbf{S}}

\newtheorem{theorem}{Theorem}
\theoremstyle{definition}
\newtheorem{definition}{Definition}[section]

\begin{document}

\maketitle

\begin{abstract}
Using observational data to learn causal relationships is essential when randomized experiments are not possible, such as in healthcare. Discovering causal relationships in time-series health data is even more challenging when relationships change over the course of a disease, such as medications that are most effective early on or for individuals with severe disease. Stage variables such as weeks of pregnancy, disease stages, or biomarkers like HbA1c, can influence what causal relationships are true for a patient. However, causal inference within each stage is often not possible due to limited amounts of data, and combining all data risks incorrect or missed inferences. To address this, we propose Causal Discovery with Stage Variables (CDSV), which uses stage variables to reweight data from multiple time-series while accounting for different causal relationships in each stage. In simulated data, CDSV discovers more causes with fewer false discoveries compared to baselines, in eICU it has a lower FDR than baselines, and in MIMIC-III it discovers more clinically relevant causes of high blood pressure.
\end{abstract}

\section{Introduction}
\label{sec:intro}

Causal relationships often change over time. At an individual level, what motivated you to be active as a child may differ from your motivations as an adult. At a societal level, new laws change the behavior of companies and citizens. While it may seem that causal inference is impossible when the relationships to be found are a moving target, relationships often change systematically, particularly in the context of health. Many diseases (such as cancer, or chronic kidney disease - CKD) are staged, allowing patients to be grouped into clusters with similar severity and for whom similar treatments may be appropriate. Similarly, hemoglobin A1C (HbA1c) provides insight into average glucose over time, week of gestation are used to help understand the progression of pregnancy, and the Simplified Acute Physiology Score (SAPS II) is used to determine the severity of illness on admission to an intensive care unit (ICU). We refer to such variables as stage variables, defined below.

\begin{definition}[Stage variable]
A variable whose values change over time, either monotonically or not, and which may alter the underlying causal structure of the data generating process.
\end{definition}

These indicators may change monotonically over time (like week of pregnancy) or in more complex ways (as with HbA1c). Some may be determined by a medical professional (as in cancer staging), while others may be identified with laboratory tests (as with kidney function). What is common to all is that stages can change within an individual, and causal relationships may differ across stages. For example, treatments that work in one stage may not work in another, such as medications that are more effective at controlling symptoms of early stage CKD compared to later stages \citep{qaseem2013screening}.

There have been no methods for learning causal relationships that are designed for such a scenario. Yet ignoring stages means at best we will not know when a relationship is most effective (e.g., not knowing whom a medication may help) and at worst may miss a cause entirely (e.g., when it is true for only a short time). Instead, leveraging stage information can lead to more accurate causal inference with less data. We aim to learn both causal relationships in staged data and the specific stages in which they apply. Further, by using existing knowledge about how stages are related we can enable causal inference in stages where data is difficult or costly to obtain.

Learning causal relationships from observational time-series data is a widely researched area and includes methods such as Granger causality \citep{granger1969investigating}, dynamic Bayesian networks (DBNs) \citep{murphy2002dynamic}, and logic-based methods \citep{kleinberg2013causality}. While many methods assume stationarity (i.e., causal structure and/or parameters do not change over time), this is highly limiting for real-world applications. Further, while DBNs have been extended to non-stationary data \citep{husmeier2010inter,robinson2010learning}, this does not find stage-dependent relationships, and cannot share data across stages. Yet the success of interventions to treat cancer depend strongly on how advanced it is \citep{greene2008staging}. In stage I, cancer is localized and surgery has a high chance of success, whereas in stage III, the cancer has spread to the lymph nodes and treatments such as chemotherapy are more commonly pursued. While the efficacy of interventions changes systematically by stage, existing methods cannot learn the conditions where relationships apply or leverage data from other stages (if causal inference proceeds separately within each). This is critical for applications where it is difficult or expensive to collect data, and limited evidence is available within each time frame.

To address this, we introduce Causal Discovery with Stage Variables (CDSV). This approach allows inference of stage-dependent causal relationships, while providing transfer of information across stages according to a user-specified weighting function. We use prior medical knowledge of the similarity between stages to guide our transfer via a weighting function. This makes CDSV especially useful for healthcare applications where we may have many samples in one stage but few samples in another (e.g., diseases with low survival rates or that are highly treatable and rarely advance). We demonstrate that CDSV outperforms the state of the art on four simulated datasets, find that it discovers medically plausible causes of high blood pressure (BP) in the MIMIC-III ICU dataset \citep{johnson2016mimic}, and show it has a low false discovery rate when finding causes of low oxygen saturation in the eICU dataset \citep{pollard2018eicu}.

Our key contributions are:
\begin{itemize}
    \item A novel method for causal discovery in staged time-series data with stage dependent learned relationships across stages.
    \item Empirical demonstration of the need to capture staging information to reduce false positives in simulated and real-world health data.
    \item Application to real-world MIMIC-III and eICU data to discover causes of high blood pressure and low oxygen saturation, respectively, for patients with varying ICU severity scores.
\end{itemize}

\section{Related Work}
We survey related work in three key areas: causal inference from observational data, causal transportability, and stage aware prediction.

\subsection{Causal Discovery}
Many approaches for learning causal models are based on Bayesian networks (BNs) \citep{pearl2000models}, directed acyclic graphs in which nodes represent variables and edges indicate causal relationships. The PC algorithm \citep{spirtes1991algorithm} is a widely used method for learning BNs from data and works well for high dimensional data with sparse relationships \citep{kalisch2007estimating}. It has been extended to time-series data via PCMCI \citep{runge2019detecting} and PCMCI+ \citep{runge2020discovering}, which adds support for contemporaneous relationships. FCI \citep{spirtes2000causation} and its temporal extension, tsFCI \citep{entner2010causal}, similarly use repeated conditional independence tests to identify causal structure, but unlike PCMCI, they allow for latent variables. CD-NOD \citep{huang2020causal} combines the PC algorithm with non-parametric kernels for causal discovery in non-stationary data. This method assumes causal structure changes are driven by a hidden variable and also uses a kernel-based conditional independence test \citep{zhang2012kernel} which has cubic computational complexity in the number of time steps. While these methods provide theoretical guarantees for causal discovery, they do not account for stages explicitly and most assume stationarity. Thus they will find incorrect causal relationships or strengths.

BNs have also been extended to time series data through DBNs \citep{murphy2002dynamic}, which use a series of BNs connected across time to indicate how variables at one time influence those at a later time. However, exact inference in BNs and DBNs is intractable and approximate methods such as MCMC sampling and variational inference must be used. These methods assume the underlying causal structure remains fixed, which is not true for staged data. One could include stage as a variable, but this may violate the assumptions made. Non-stationary DBNs \citep{robinson2008non} allow for causal relationships that change over time and have been extended to support continuous data, latent variables, and changepoint detection \citep{grzegorczyk2009non,shafiee2020non}. While these methods allow for changing causal structure over time, the transitions between the graphs are limited: single edges are added/deleted or two graphs are merged. In contrast, we allow arbitrary changes in causal structure between stages with exact inference.

Other methods for causal inference in time series include Granger causality and logic-based methods.
If past values of $X$ improve prediction of future values of $Y$, $X$ is said to Granger-cause $Y$ \citep{granger1969investigating}. 
Granger causality has been extended to multivariate data and to support non-linear relationships \citep{ancona2004radial,marinazzo2011nonlinear}. However, it may miss causes in stages with less data if inference is done within each stage, and may have both false positives and negatives when combining data across stages. 
Finally, \citet{kleinberg2013causality} introduced a method that represents causal relationships as probabilistic computation tree logic formulas \citep{hansson1994logic}, and uses measures of causal significance to distinguish causation from correlation. This allows complex relationships such as variables true for a duration of time, or a conjunction of factors causing an effect. The approach uses a time window for each relationship (i.e. $x$ causes $y$ in 10-20 time units) rather than a single time lag and this window can be discovered automatically.
While this method does not allow for nonstationary relationships, we extend it to build CDSV and discuss it further in methods.

\subsection{Causal Transport}
Another related area is causal transportability, which defines the theoretical conditions under which causal relationships in one dataset can be applied to another. For data with stage variables, one could split the data into separate datasets for each stage and use the causal transportability methods to infer causal relationships across stages. The primary work in this area \citep{bareinboim2013general,bareinboim2014transportability,lee2013m} defines the necessary and sufficient conditions where treatment effects from one dataset can be applied to another. \citet{pearl2014external} introduced selection variables, which complement a structural causal model (SCM) and describe how its equations change between different populations. Using the SCM on different populations requires conditioning on different values of these selection variables. Specifying these variables requires the full SCM to be specified, which requires prior knowledge of which variables will change in different datasets.
Further, these methods are not directly applicable to time-series data. 
Instead, we propose a method that only requires the stage variable to be specified without requiring knowledge of which variables and relationships will change across stages.

\subsection{Staged Health Prediction}
While stage variables have not been used in causal inference, StageNet \citep{10.1145/3366423.3380136} incorporated them into predictive models of health risk and for patient subtyping on the MIMIC-III and End-Stage Renal Disease datasets. This deep learning based architecture uses an LSTM with a stage-aware convolutional module to predict patient mortality risk. While this model has good prediction performance compared to baselines, it does not provide insight into causes of the predicted outcome, and requires a large number of samples, whereas in our setting some stages may have far less data (e.g., late stage cancer).

\section{Methods}

\begin{figure}[t]
 % \floatconts
  \label{fig:timeline}
  {\includegraphics[width=0.9\columnwidth]{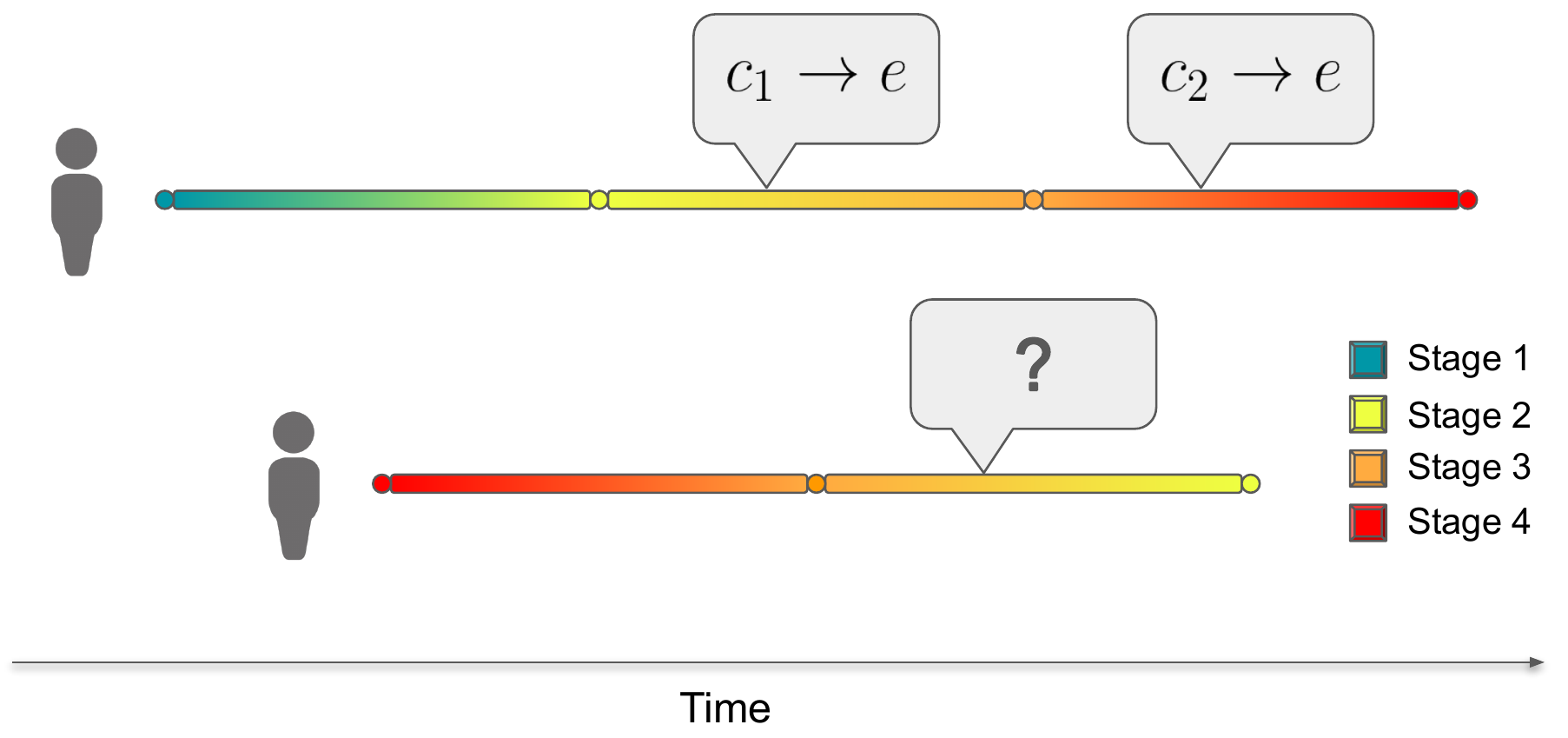}}
    {\caption{Example of stage based causal inference with transfer. Each person start and ends in different stages. From the top person we are confident in the causal relationships learned in stages 2 and 3, so CDSV uses a weighting function to transfer this knowledge to the second person for inference in stage 3. Circles indicate stage transitions.}}
\end{figure}

We introduce a new method for causal discovery in time-series data called Causal Discovery with Stage Variables (CDSV). CDSV allows data from multiple stages to be used for learning causal structure, while still enabling inference of stage-dependent relationships (e.g., learning different causes of diet early in pregnancy versus later on). We do this by weighting data from different stages during causal inference for each target stage. This allows information to be transferred between stages, which can help overcome noise (e.g., when stages are fuzzy or uncertain), while providing stage-dependent output. 

\subsection{Problem setup}
We begin with multivariate data $\bX$ consisting of a variable length time-series in our observations $\mathbb{R}^{p \times d_p \times t_p}$ where $d_p$ is the number of variables and $t_p$ is the number of time steps for time-series $p$. Additionally we assume we are given stage variables, $\bS$, for each patient. A person may be in the same stage for their entire time series or may change over time.

%Stage variables are pervasive as we have noted. People with chronic health conditions (such as diabetes) are regularly monitored and progressive illnesses such as multiple sclerosis and ALS have associated scores assessing disability and disease progression. Many other stages (such as weeks of gestation or risk scores in ICU patients) are currently available but not frequently used in causal inference. 

\subsection{Kleinberg Method}

We then aim to determine the causes of a set of effects (outcomes) in this time-series. See Figure \ref{fig:timeline} for an illustration of this setup. To do this we extend the method of \citet{kleinberg2013causality}. In this approach a causal relationship is represented by the Probabilistic Computation Tree Logic (PCTL) \citep{hansson1994logic} formula:\footnote{PCTL was augmented with a lower time bound on leads-to formulas by \citet{kleinberg2013causality}. We use that version here.}
\begin{equation}
    c \rightsquigarrow^{\ge k, \le l}_m e.
\end{equation}
This formula means that cause $c$ leads to the effect $e$ occurring in $k$ to $l$ time units, with probability $m$.

For example, consider the problem of determining the effects of different treatments on the outcome of cancer remission for patients in the 4 different stages of cancer. Each patient $p$ would have their own time-series with the outcomes of cancer remission or progression. Each time-series has a variable number of data points as each patient may begin treatment at a different time and each will progress through different stages of cancer at different rates.

The causal significance of $c$ for $e$ is given by:
\begin{equation}
    \epsilon_{avg}(c, e) =
    \frac{\sum_{x \in X} P(e|c \land x) - P(e|\neg c \land x)}{|X \setminus c|} %\label{eq2}
\end{equation}
\begin{equation}
    P(e|c \land x) = \sum_{t \in T}
    \frac{\#(c_t, x_{i:j}, e_{k:l})}
    {\#(c_t, x_{i:j})} %\label{eq3}
\end{equation}
Note that $\#$ above represents the number of times that its arguments occur in the specified time window, with $c$ occurring at time $t$, $x$ occurring in $i$ to $j$ time units before $e$, and $e$ occurring in $k$ to $l$ time units after $t$. $X$ is the set of potential causes (variables that raise the probability of $e$). By subtracting the frequency based probabilities of the cause and effect occurring versus the effect occurring without the cause over all instances in the data, this method yields the average causal significance of all potential causes. A potential cause with $\epsilon_{avg}$ greater than some threshold are considered $\epsilon$-significant causes.

This method requires two primary assumptions to guarantee the correctness of discovered causes:

\begin{enumerate}
    \itemsep 0em 
    \item The causal relationships between observed variables do not change over time.
    \item The common causes of all pairs of variables must be included in the data.
\end{enumerate}

In this paper, we extend \citet{kleinberg2013causality}  to account for stage-dependent relationships while also allowing information sharing across stages. Additionally, we reduce the stationarity requirement as relationships may change with the stage variables.

\subsection{CDSV Assumptions}
To guarantee that CDSV finds the correct causal relationships, the causal relationships in the time-series $\bX$ must be stationary within each stage. Note that this is a less strict assumption than requiring stationarity across the entire time series. A key advantage of our method is that relationships can change between stages. Second, we require causal faithfulness, meaning that the observed data follows the underlying causal structure in each stage. Third, we must have a user-defined weighting function to determine how strongly to weigh data from different stages. Lastly, we assume we have stage variable values for each point in the time-series.

\subsection{Causal Significance with Stage Variables}

Building on \citet{kleinberg2013causality} we introduce the CDSV algorithm. CDSV takes time-series data $\bX$, stage variables (such as stage of cancer, HbA1c, or SAPS II score) $\bS$, a set of variables (potential causes and effects), a minimum time lag, a maximum time lag, a significance threshold, and a weighting function $w(s,o)$ as input. The minimum and maximum time lags define the range of time windows to test for causal relationships and the weighting function defines how stages are related for transfer between stages. The significance threshold defines how large the causal significance must be to consider it a true causal relationship. CDSV returns the set of significant causes and effects in the time-series along with time windows and the strength of each cause.

CDSV has two steps. First it examines each cause and effect pair within each stage, holding out other potential causes and computing an average difference in probability of the effect given the cause and each other potential cause. This step is the same as in \citet{kleinberg2013causality}. CDSV then multiplies the resulting average difference by the weighting function $w(s, o)$, which returns the weight between stage $s$ and stage $o$, to get the causal significance for the given causal relationship.

\subsection{Inference and Computational Complexity}

We compute the probabilities in our method by summing the number of relevant events in the appropriate time window as defined by the equation below. Here the time window is defined by $c$ occurring at time $t$, $x$ occurring in $i$ to $j$ time units before $e$, and $e$ occurring in $k$ to $l$ time units after $t$.
\begin{equation}
    P_o(e|c \land x) = \sum_{t \in T}
    \frac{\#_o(c_t, x_{i:j}, e_{k:l})}
    {\#_o(c_t, x_{i:j})} \label{eq3}
\end{equation}

Equation \ref{eq2} below defines the core of our computation and Algorithm \ref{alg:cdsv} defines the step by step method.
\begin{equation}
    \small
    \epsilon_{avg}(c, e, s) = \sum_{o \in S}
    \frac{\sum\limits_{x \in X} P_o(e|c \land x) - P_o(e|\neg c \land x)}{|X \setminus c|} \cdot w(s, o) \label{eq2}
\end{equation}
Equation \eqref{eq2} gives the significance of cause $c$ for effect $e$ within stage $s$. Here $X$ is the full set of potential causes of $e$ in $s$ (i.e., all variables that raise the probability of $e$). Both $c$ and all other $x \in X$ have associated causal relationships of the form $x \leadsto^{\geq i, \leq j} e$.

\begin{algorithm}[tb]
\caption{Causal Significance with Stage Variable}
\label{alg:cdsv}
\begin{algorithmic}[1] %[1] enables line numbers
\FOR{ $c \rightsquigarrow^{\ge k, \le l} e$ for patient $p$ in stage $s$}
\FOR{every stage $o \in S$}
\STATE $\epsilon_o(c, e) = $ Causal significance of $c$ for $e$ in stage $o$
\STATE $\epsilon_{avg}(c, e, s) += \epsilon_o(c, e) \cdot w(s,o)$
\ENDFOR
\IF{$\epsilon_{avg}(c, e, s) \ge \epsilon$}
\STATE $R[c, e, s] = \epsilon_{avg}(c, e, s)$
\ENDIF
\ENDFOR
\STATE \textbf{return} $R$
\end{algorithmic}
\end{algorithm}
We use two weighting functions in our experiments:

\begin{equation}
    w(s, o) = 1 - \frac{|s-o|}{T}
    \label{eq:linear-weight}
\end{equation}
\begin{equation}
    w(s, o) = 
    \begin{cases}
        \left( 1 - \frac{|s-o|}{T} \right) \cdot \frac{N_o}{N_s} & \text{if } N_s \le \tau \\
        \ \ 1 & \text{if } N_s > \tau, s = o \\
        \ \ 0 & \text{if } N_s > \tau, s \ne o
    \end{cases}
    \label{eq:sample-weight}
\end{equation}

Weighting function \ref{eq:linear-weight} linearly weights data from other stages based on the absolute difference between the stages. Here $T$ is the number of stages and data from other stages $o$ is weighted lower the further away it is from the current stage $s$. For example, with 3 stages, $w(1,1) = 1, w(1,2) = 2/3, \text{and } w(1,3) = 1/3$.

Function \ref{eq:sample-weight} is similar to function \ref{eq:linear-weight} but multiplies by the additional $\frac{N_o}{N_s}$ ratio which is the number of samples in stage $o$ divided by the number of samples in stage $s$. Multiplying by this ratio ensures that stages with more samples will have more weight than those with fewer. Additionally, we define a threshold $\tau$ which collapses the function to either 1 or 0 once a stage $s$ has $\tau$ data points. This guarantees that with more than $\tau$ samples in stage $s$, only data from that stage is used. For a proof of data consistency in the infinite data limit, see Appendix \ref{apd:data-consistent}.

Note that the above weighting functions were appropriate for our experiments, but any weighting function may be used in our method. We recommend domain experts determine the appropriate weighting function for different problems in healthcare. For additional information on how to determine weighting functions see Appendix \ref{apd:weighting-func}.

The computational complexity of our method is $O(N^3T)$ with $N$ variables and $T$ being the length of the time series. Weighting does not increase the time complexity, so this is the same as in \citet{kleinberg2013causality}. As each relationship can be tested independently of the others, we parallelize our implementation for significant performance gains.

\subsection{Finding Significant Relationships}

In Algorithm \ref{alg:cdsv} note on line 6 that a potential causal relationship is only significant if it is greater than or equal to some threshold $\epsilon$. To choose this threshold in a principled way, we compute p-values for each causal significance score. From \citet{kleinberg2013causality}, we know that if the set of causal significance values $R$ follows a mean 0 normal distribution, then there are no significant causal relationships found. So to compute p-values for each value in $R$, we fit a normal distribution to this data. With this empirical null distribution, we calculate the p-values of each causal relationship and then control for multiple hypothesis testing using the Benjamini-Hochberg correction at an FDR level of 0.05.

\section{Experiments}
We run experiments on six datasets to evaluate CDSV. First we evaluate on four simulated datasets: a simple structure with three stages and three variables, another with two stages and three variables where a cause switches from positive to negative, a complex structure with three stages and 10 variables, and lastly the same complex structure with few samples in stage 1. On simulated data, we compare our approach to state of the art methods (tsFCI, PCMCI and PCMCI+) using area under the receiver operating characteristic curve (AUROC), false positive rate (FPR) and false negative rate (FNR), comparing inferences to the true causal relationships generated and their true timing. We do not compare to CD-NOD due to its cubic time complexity, which is not appropriate for our long time-series data. As a real-world test, we use two large hospital and ICU datasets, eICU and MIMIC-III. On MIMIC-III we compare CDSV against the baselines to find causes of high blood pressure and on eICU we aim to find causes of low oxygen saturation and compare to PCMCI and PCMCI+.

\subsection{Simulated Data}

\subsubsection{Simple Simulation}

\begin{table*}
\setlength{\tabcolsep}{4pt}
\centering
\begin{tabular}{lrrr|rrr|rrr|rrr}
\toprule
& \multicolumn{3}{c}{Simple} & \multicolumn{3}{c}{Flipped relationship} &
\multicolumn{3}{c}{Complex} &
\multicolumn{3}{c}{Complex Small Sample} \\
Method & AUC & FPR & FNR & AUC & FPR & FNR & AUC & FPR & FNR & AUC & FPR & FNR \\
\midrule
\citet{kleinberg2013causality}     & 0.57 & 0.16 & 1.00 & \textbf{1.00} & \textbf{0.00} & \textbf{0.00} & 0.52 & 0.07 & 0.80 & 0.00 & 1.00 & 1.00       \\
CDSV (this work)           & \textbf{1.00}  & \textbf{0.00} & \textbf{0.00} & \textbf{1.00} & \textbf{0.00} & \textbf{0.00}  & \textbf{0.55} & \textbf{0.02} & 0.74 & \textbf{0.55} & \textbf{0.02} & \textbf{0.78}       \\
PCMCI           & 0.50  & 0.67 & 1.00 & 0.29 & 0.57 & 1.00 & 0.00 & 1.00 & 1.00 & 0.00 & 1.00 & 1.00      \\
PCMCI+          & 0.50  & 0.67 & 1.00 & 0.29 & 0.57 & 1.00 & 0.40 & 0.75 & \textbf{0.61} & 0.00 & 1.00 & 1.00      \\
tsFCI           & 0.50  & \textbf{0.00} & 1.00 & 0.50 & \textbf{0.00} & 1.00 & 0.49 & 0.15 & 1.00 & 0.00 & 1.00 & 1.00      \\
\bottomrule
\end{tabular}
\caption{Results on simulated data. AUC is area under the receiver operating characteristic curve, and a value of 0.00 means that method found no valid causal relationships. \citet{kleinberg2013causality} is the base method, inferring causes within each stage separately. The best value in each column is indicated in bold. Note that we consider the relationship and its timing in evaluation.}
\label{tab:allres}
\end{table*}

This dataset tests the case where a causal relationship weakens as stages progress (stage 1 to stage 2) and eventually ceases (stage 3). 
This scenario is found in real world cases such as CKD, where early stages of the disease are treated with medication to relieve swelling and control blood pressure but medications become less effective as the disease progresses. In stage 5, CKD medications are no longer effective and dialysis or kidney transplant are required.

We generate a three stage dataset with 3 binary variables: $x_1, x_2,$ and $y$. We generate 5 time-series in each stage resulting in 15 total time-series with 1000 time points in each. In this data, $x_2$ is a noise variable that does not interact with other variables. In stage 1, $x_1$ causes $y$ in two time steps with probability 0.9. In stage 2 the probability is reduced to 0.52. Finally, in stage 3 $x_1$ does not cause $y$ at all, and $y$ occurs spontaneously with probability 0.5.

We evaluate CDSV using linear weighting of stages compared to the base method of \citet{kleinberg2013causality}, where we do causal inference within each stage without sharing information across stages. We expect that CDSV will perform best as it can borrow data from the first stage when doing inference in the second stage where the causal relationship is weak (i.e., using evidence from stage 1 to augment the observations in stage 2). We compare against tsFCI, PCMCI, and PCMCI+. These methods were selected as they are the closest to our approach, and support time series data. For fair comparison with CDSV, we provide the stage variable to all baselines as another input.

As shown in Table \ref{tab:allres}, tsFCI, PCMCI and PCMCI+ perform poorly in comparison to CDSV. When using an FDR of 0.05 both PCMCI baselines find the incorrect causal relationship of $x_1$ causes $y$ in 3 time steps. They also find that each stage causes $y$, which is incorrect. tsFCI has a lower FPR than the PC methods, but still has a high FNR. As a further baseline, we use the causal inference method of \citet{kleinberg2013causality}, conducting causal inference using data from each stage independently. This method does slightly better than the baselines at an FDR of 0.05, finding correctly that $x_1$ causes $y$ in 2 time steps in stage 1 for an AUROC of 0.57, however this method fails to find that $x_1$ causes $y$ in stage 2, even though this is a valid causal relationship. CDSV, \citet{kleinberg2013causality}, and tsFCI all have low FPR in comparison to PCMCI and PCMCI+ but only CDSV has both 0 FPR and FNR. These methods are more conservative when selecting a potential relationship as true in comparison to PCMCI+. tsFCI is conservative because it assumes latent variables may exist and therefore requires strong conditional independence in the data.

CDSV performs best as it discovers both correct causal relationships for stage 1 and 2 at an FDR of 0.05, and does not find any erroneous causal relationships in stage 3. These results confirm our hypothesis that CDSV performs best by reweighting the data from stage 1 such that CDSV is able to learn the weaker causal relationship in stage 2 more easily.

\subsubsection{Flipped Two Stage Simulation}
We now examine performance when a cause switches from positive to negative across stages. For example, metformin is a common early treatment for diabetes, but can decrease or increase blood glucose in people with diabetes based on the stage of their disease \citep{gormsen2019metformin}. To test this case we generate a two stage dataset with 3 binary variables: $x_1$, $x_2$, and $y$. In stage 1, $x_1$ causes $y$ to be 1 with probability 0.9 and in stage 2 $x_1$ causes $y$ to be 0 with probability 0.9. We again have $x_2$ as a noise variable which does not affect others. We generate 5 time-series in each stage with 1000 time points each.

As shown in Table \ref{tab:allres}, CDSV and \citet{kleinberg2013causality} both correctly discover the stage-dependent relationships, while other methods find incorrect relationships. PCMCI+ and tsFCI do not find any correct causal relationships because they treat the data equivalently across stages, which results in stage 1 and stage 2 cancelling out because of the flipped relationship. Since the \citet{kleinberg2013causality} method infers causes in each stage separately, it is also able to find the correct relationships. Despite the information sharing used by CDSV, it still finds the correct relationships because it puts the most weight on data from the same stage.

%Since the \citet{kleinberg2013causality} method also infers causes while ignoring stage, it also finds incorrect relationships. Note that \citet{kleinberg2013causality} and tsFCI both have low FPR but high FNR as they are more conservative than PCMCI.

\subsubsection{Complex Three Stage Simulation}

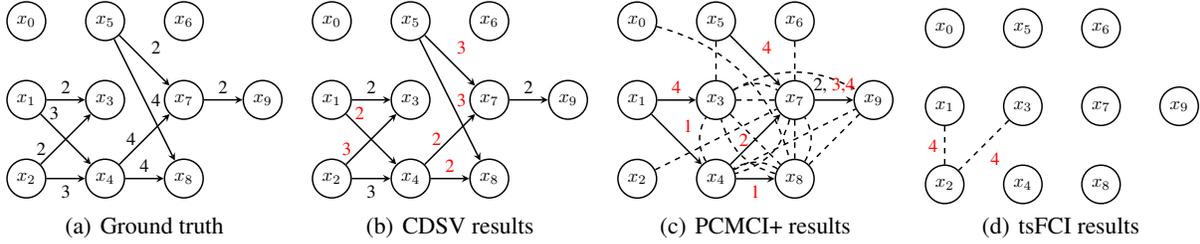
\begin{figure*}[t]
%    \floatconts
    \label{fig:complex-experiment}
    {
        \subfigure[Ground truth]{
            \resizebox{0.23\textwidth}{!}{
            \begin{tikzpicture}[> = stealth, auto, node distance = 1.5cm, thick]
                \tikzstyle{every state}=[
                    draw = black,
                    thick,
                    fill = white,
                    minimum size = 4mm
                ]
                \node[state] (0) {$x_0$};
                \node[state] (1) [below of=0] {$x_1$};
                \node[state] (2) [below of=1] {$x_2$};
                \node[state] (3) [right of=1] {$x_3$};
                \node[state] (4) [right of=2] {$x_4$};
                \node[state] (5) [right of=0] {$x_5$};
                \node[state] (6) [right of=5] {$x_6$};
                \node[state] (7) [right of=3] {$x_7$};
                \node[state] (8) [right of=4] {$x_8$};
                \node[state] (9) [right of=7] {$x_9$};
                \path[->] (1) edge node {2} (3);
                \path[->] (1) edge node[above=14pt, left=0pt] {3} (4);
                \path[->] (2) edge node[below=5pt, left=7pt] {2} (3);
                \path[->] (2) edge node[below] {3} (4);
                \path[->] (4) edge node[above=0pt, left=1pt] {4} (7);
                \path[->] (4) edge node {4} (8);
                \path[->] (5) edge node {2} (7);
                \path[->] (5) edge node[right=0pt] {4} (8);
                \path[->] (7) edge node {2} (9);
            \end{tikzpicture}
            }
            \label{fig:complexSimStruct}
        }
        \subfigure[CDSV results]{
            \resizebox{0.23\textwidth}{!}{
            \begin{tikzpicture}[> = stealth, auto, node distance = 1.5cm, thick]
                \tikzstyle{every state}=[
                    draw = black,
                    thick,
                    fill = white,
                    minimum size = 4mm
                ]
                \node[state] (0) {$x_0$};
                \node[state] (1) [below of=0] {$x_1$};
                \node[state] (2) [below of=1] {$x_2$};
                \node[state] (3) [right of=1] {$x_3$};
                \node[state] (4) [right of=2] {$x_4$};
                \node[state] (5) [right of=0] {$x_5$};
                \node[state] (6) [right of=5] {$x_6$};
                \node[state] (7) [right of=3] {$x_7$};
                \node[state] (8) [right of=4] {$x_8$};
                \node[state] (9) [right of=7] {$x_9$};
                \path[->] (1) edge node {2} (3);
                \path[->] (1) edge node[above=14pt, left=0pt] {\color{red} 2} (4);
                \path[->] (2) edge node[below=5pt, left=7pt] {\color{red} 3} (3);
                \path[->] (2) edge node[below] {3} (4);
                \path[->] (4) edge node[above=0pt, left=1pt] {\color{red} 2} (7);
                \path[->] (4) edge node {\color{red} 2} (8);
                \path[->] (5) edge node {\color{red} 3} (7);
                \path[->] (5) edge node[right=0pt] {\color{red} 3} (8);
                \path[->] (7) edge node {2} (9);
            \end{tikzpicture}
            }
            %\caption{Results from CDSV %The edge weights represent the different timings of relationships. The red colored edges are incorrect, but only in terms of timing, notice that all the edges match the ground truth diagram in structure.
            %}
            \label{fig:complexSimCDSV}
        }
        \subfigure[PCMCI+ results]{
            \resizebox{0.23\textwidth}{!}{
            \begin{tikzpicture}[> = stealth, auto, node distance = 1.5cm, thick]
                \tikzstyle{every state}=[
                    draw = black,
                    thick,
                    fill = white,
                    minimum size = 4mm
                ]
                \node[state] (0) {$x_0$};
                \node[state] (1) [below of=0] {$x_1$};
                \node[state] (2) [below of=1] {$x_2$};
                \node[state] (3) [right of=1] {$x_3$};
                \node[state] (4) [right of=2] {$x_4$};
                \node[state] (5) [right of=0] {$x_5$};
                \node[state] (6) [right of=5] {$x_6$};
                \node[state] (7) [right of=3] {$x_7$};
                \node[state] (8) [right of=4] {$x_8$};
                \node[state] (9) [right of=7] {$x_9$};
                \path[dashed] (0) edge[bend left] node {} (8);
                \path[->] (1) edge node {\color{red} 4} (3);
                \path[->] (1) edge node {\color{red} 1} (4);
                \path[dashed] (2) edge node {} (7);
                \path[dashed] (3) edge[bend left] node {} (7);
                \path[dashed] (3) edge node {} (8);
                \path[dashed] (3) edge[bend left] node {} (9);
                \path[dashed] (4) edge[bend left] node {} (3);
                \path[->] (4) edge node[left] {\color{red} 2} (7);
                \path[->] (4) edge node[below] {\color{red} 1} (8);
                \path[dashed] (4) edge node {} (9);
                \path[dashed] (5) edge node {} (3);
                \path[->] (5) edge node {\color{red} 4} (7);
                \path[dashed] (6) edge node {} (7);
                \path[dashed] (7) edge node {} (3);
                \path[dashed] (7) edge[bend left] node {} (4);
                \path[dashed] (7) edge node {} (8);
                \path[->] (7) edge node {2, \color{red}3,4} (9);
                \path[dashed] (8) edge[bend left] node {} (3);
                \path[dashed] (8) edge[bend right] node {} (4);
                \path[dashed] (8) edge[bend right] node {} (7);
                \path[dashed] (8) edge node {} (9);
            \end{tikzpicture}
            }
            \label{fig:complexSimPCMCI}
        }
        \subfigure[tsFCI results]{
            \resizebox{0.23\textwidth}{!}{
            \begin{tikzpicture}[> = stealth, auto, node distance = 1.5cm, thick]
                \tikzstyle{every state}=[
                    draw = black,
                    thick,
                    fill = white,
                    minimum size = 4mm
                ]
                \node[state] (0) {$x_0$};
                \node[state] (1) [below of=0] {$x_1$};
                \node[state] (2) [below of=1] {$x_2$};
                \node[state] (3) [right of=1] {$x_3$};
                \node[state] (4) [right of=2] {$x_4$};
                \node[state] (5) [right of=0] {$x_5$};
                \node[state] (6) [right of=5] {$x_6$};
                \node[state] (7) [right of=3] {$x_7$};
                \node[state] (8) [right of=4] {$x_8$};
                \node[state] (9) [right of=7] {$x_9$};
                \path[dashed] (2) edge node {\color{red} 4} (1);
                \path[dashed] (3) edge node {\color{red} 4} (2);
            \end{tikzpicture}
            }
            \label{fig:complexSimTsfci}
        }
    }
       {\caption{Ground truth (a) and inferred causal relationships (b, c, d) for our complex (10 variable) simulation. Annotations on edges indicate timing of each relationship. Solid edges indicate correct relationships, dashed edges are incorrect. Red numbers along edges indicate an incorrectly inferred timing (though a true relationship exists between the nodes). Notably, all errors by CDSV are in timing, while all relationships are recovered with no false positives.}}
\end{figure*}

To test how our approach scales, we generate another three stage dataset, now with 10 variables and more causal relationships among them. We create 3 layers of 3 variables with a final layer containing a single outcome variable. We randomize the causal edges between these layers and the time lags of each relationship. The structure is shown in Figure \ref{fig:complexSimStruct}. The weights of causal relationships in each stage are randomized. Stage 1 causal weights are randomized between [1, 2], weights in stage 2 are between [0.1, 0.5], and the weights in stage 3 are between [0.01, .05],  replicating relationships that decrease in strength over time. We translate these weights to probabilities when forward generating the time-series by using the logistic function $\sigma(x) = 1/(1+e^{-x})$ which maps all inputs to the [0, 1] probability space. Unlike Experiment 1, relationships are present in stage 3 but are now very weak. Again we make the causal relationships strongest in stage 1, weaker in stage 2, and weakest in stage 3. To further increase complexity, we introduce more varied timing among the relationships, with randomly generated windows in $[1,4]$. We generate 5 time-series in each stage of 1000 time steps each, resulting in a final set of 15 time-series.

The AUROC, FPR, and FNR results in Table \ref{tab:allres} demonstrate that the weak relationships, extra timing complexity, and more causal relationships in this structure make it more challenging for all methods. Nevertheless, CDSV outperforms the baselines. PCMCI does not find any correct relationships at the FDR level of 0.05. One reason PCMCI performs poorly is that it treats stage variables as causes, which is incorrect in our problem setting, and the PCMCI assumption of stationarity is violated because causal relationships change in each stage. Our method is more successful because it allows causal relationships to change across each stage while transferring learned relationships between stages. We also experimented with PCMCI+ without stage variables. While this avoids finding stages as causes, it still identifies many spurious relationships. 

Figure \ref{fig:complex-experiment} shows the true structure, CDSV output, PCMCI+ output without stage information, and tsFCI output. CDSV finds all correct edges and only has small errors in the timing of some relationships. We evaluate AUROC strictly, calling such relationships false discoveries, but notably CDSV recovers relationships between all actual causes and effects with no incorrect edges. PCMCI+ without stage information instead finds many false relationships and fails to recover true relationships (such as $x_2 \rightarrow x_3$). This may be because PCMCI+ combines data across stages, finding a larger proportion of weak causal relationships resulting in the discovery of incorrect causal edges. tsFCI is more conservative than PCMCI+, but at the expense of finding no correct relationships. On FPR, CDSV has the lowest value while \citet{kleinberg2013causality} and tsFCI also have a low FPR. These methods are more conservative when selecting a potential relationship as true in comparison to PCMCI+. Note that CDSV, \citet{kleinberg2013causality}, and tsFCI have higher FNR than PCMCI+ as this is a common statistical tradeoff when a method has lower FPR.

\subsubsection{Complex Small Sample Simulation}

We now test the case of many samples in one stage and few in another stage. This scenario is prevalent in healthcare, especially when data is rare, costly, or difficult to collect. For example, cancer data will have fewer people with stage IV cancer than those in stage I because of the low life expectancy of those with stage IV cancer.

We generate a dataset with the same structure as Figure \ref{fig:complex-experiment} with one time-series in stage 1, five time-series in each of stages 2, and 3. We generate 1000 time points for each time-series and again have strong causal relationships in stage 1 with strength decreasing in each stage. In this experiment we use function \ref{eq:sample-weight} as a data consistent weighting function which gives data in stages with more samples more weight.

As shown in Table \ref{tab:allres}, only CDSV finds any true relationships in this data at an FDR level of 0.05. This demonstrates a key advantage of CDSV: leveraging information across stages to overcome a small sample size within a stage.

\subsection{Causes of low oxygen saturation in eICU}
\subsubsection{Data}
We next evaluate CDSV, PCMCI, and PCMCI+ on the eICU dataset \citep{pollard2018eicu}, a multi-center database consisting of over 200,000 ICU admissions. We create hourly time-series data for each ICU patient with stays of 1 -- 10 days. We aim to find causes of low oxygen (O$_2$) saturation with input variables of vasopressors, oxygenation treatment, heart rate, respiratory rate, and others. We use the Acute Physiology and Chronic Health Evaluation (APACHE) IV score \citep{zimmerman2006acute} as our stage variable. APACHE IV is a medically tested scoring system that assesses the severity of ICU patients and is available in the eICU dataset. We categorize patients into two APACHE score ranges: low ($0 - 70$) and high ($\ge 70)$. A higher APACHE score means the patient is in worse condition and we again use empirical quantiles to define our score categories. Our final data consists of 70,151 patients with the longest time-series having 579 hours and the shortest having 24 hours. See Appendix \ref{apd:eicu-experiment} for experiment details.

\subsubsection{Results}

CDSV finds no significant causes of low O$_2$ saturation when testing relationships between 1 and 4 hours with an FDR level of $0.05$. While we hypothesized that oxygenation treatment would have a negative effect on low O$_2$ saturation in the future, CDSV was not confident in this cause due to its low frequency in data. When ranking causal significance without thresholding p-values in CDSV, we find that normal O$_2$ saturation has a negative effect on low O$_2$ for 1 to 4 hours in both APACHE score stages.

PCMCI and PCMCI+ find 65 and 62 causes respectively of low O$_2$ saturation at an FDR of 0.05. Most of these causes are implausible, but both baselines find that normal O$_2$ saturation has a negative effect on low O$_2$ 1--4 hours in the future.

While normal O$_2$ does not directly prevent low O$_2$ saturation, we see this result because the data contains many patients with normal O$_2$ who never experience low O$_2$ saturation. Due to this, CDSV and the baselines both infer that normal O$_2$ has a negative effect on low O$_2$, meaning normal O$_2$ saturation will lower the chance of low O$_2$ occurring in 1-4 hours. However, CDSV correctly finds that this cause is not significant using FDR correction while both baselines find this is a significant cause, which is incorrect. This demonstrates that CDSV is more conservative than other methods at the same FDR level.

\subsection{Causes of high blood pressure in MIMIC-III}
\subsubsection{Data}
As a second real-world evaluation, we test all methods on MIMIC-III \citep{johnson2016mimic}, a publicly available critical care dataset. In this experiment we aim  to find causes of high blood pressure (BP) given the input variables of mechanical ventilation, vasopressors, colloid bolus, crystalloid bolus, non-invasive ventilation, heart rate, and respiratory rate. We select these variables because they directly affect blood pressure and have low missingness in MIMIC-III. We use the Simplified Acute Physiology Score (SAPS-II) \citep{le1993new} as our stage variable. This score is a medically validated severity score for patients in the ICU that combines 15 variables and is calculated on admission. We categorize each patient into 3 stages: low SAPS-II score (0-22), medium SAPS-II score (22-48), and high SAPS-II score (48-163). A higher SAPS-II score means the patient is in worse condition. The ranges were determined using the empirical quantiles of the distribution of scores for every patient.

The MIMIC-Extract project \citep{wang2020mimic} provides hourly laboratory measurements, vital signs, and interventions including vasopressors and ventilation.
%This project provides unit conversions between different measurement types, outlier removal, and grouping of similar features to reduce unnecessary redundancy in the raw MIMIC data.
The MIMIC Code Repository project \citep{johnson2018mimic} similarly provides code to standardize concepts such as severity scores. For this experiment, we combine the SAPS-II score and the projects above. The final extracted data has 28,283 patients with 29 variables each.
The longest time-series has 240 hours worth of data and the shortest time-series is 24 hours.

%Given that we have hourly data for vasopressors, ventilation, boluses, and vitals
We choose the well-understood outcome of high BP to compare CDSV to baselines. We hypothesize that vasopressors and boluses should increase BP in 1 to 4 hours. Using SAPS-II as our stage variable and labs, vitals as our data along with high BP as our outcome, we apply CDSV and the baselines to learn causal relationships. See Appendix \ref{apd:mimic-experiment} for experimental details.

\begin{table}
\centering
\setlength{\tabcolsep}{3pt}
\begin{tabular}{lrr}
\toprule
Causal Relationship & Time & Stage \\
\midrule
Colloid Bolus  & 2 to 4 hours & Low SAPS score  \\
Low heart rate & 1 to 2 hours & Low SAPS score \\
\bottomrule
\end{tabular}
\caption{Causes found by CDSV leading to high BP on MIMIC data at FDR=0.05}
\label{tab:cdsvmimiccauses}
\end{table}

\begin{table}
\centering
\setlength{\tabcolsep}{3pt}
\begin{tabular}{lrr}
\toprule
Causal Relationship  & Time & Influence/Validity \\
\midrule
High heart rate  & 1-4 hours & + \cmark \\
Low SAPS score & 1-4 hours & + \xmark \\
High respiratory rate & 1-3 hours & + \cmark \\
Normal O2 saturation & 1-4 hours & + \xmark \\
Vasopressors & 1-4 hours & - \xmark \\
Phenylephrine & 1-4 hours & - \xmark \\
Norepinephrine & 1-4 hours & - \xmark \\
\bottomrule
\end{tabular}
\caption{Causes found by PCMCI and PCMCI+ leading to high BP on MIMIC data at FDR=0.05. Negative relationships lead to normal or lower blood pressure.}
\label{tab:pcmcimimiccauses}
\end{table}

\subsubsection{Results}
CDSV finds two causes of high BP at the FDR level of 0.05 shown in Table \ref{tab:cdsvmimiccauses}. The first cause found is that colloid boluses cause high BP in 2 to 4 hours. A colloid bolus is a liquid solution that is administered when a patient loses too much fluid in their circulatory system. The bolus increases fluid volume, which increases BP and brings the body back to homeostasis. This relationship is well supported in medical literature \citep{lewis2018colloids}. The second relationship identified, low heart rate leading to high blood pressure, may be found when a patient is prescribed beta blockers \citep{rothwell2010effects} or has thickened heart tissue \citep{karam1989hypertensive}.

In contrast, tsFCI found no causes of high BP. It found only that administration of epinephrine causes future epinephrine to be given after 4 hours. It finds no other causes or effects in the data. Since tsFCI assumes hidden confounders may exist, if a relationship can be explained by a shared parent, it is not returned as a valid relationship. MIMIC has many variables that are conditionally independent, so tsFCI does not find any causes.

Both PCMCI and PCMCI+ identified 88 causes at an FDR level of 0.05. The top scoring causes found by both algorithms are shown in Table \ref{tab:pcmcimimiccauses}. While some causes are plausible, there are many incorrect causes such as vasopressors having a negative influence on high BP, which is not supported by medical literature. Additionally, because the baselines treat the stage variable as a possible cause, they find that low SAPS score causes high BP which is not medically plausible.

PCMCI and PCMCI+ treating stage variables as potential causes leads to these methods assigning too much causal strength to many potentially invalid relationships, even at the FDR threshold 0.05. CDSV on the other hand is more conservative at the same FDR level and finds a smaller but more medically justified set of causes. CDSV also does not assume causal relationships are stationary across stages and can learn different causal relationships in each stage. 

\section{Limitations}
The main limitation of CDSV is the need for an accurate user-defined stage variable. While these exist in many datasets such as MIMIC, with progressive or graded conditions (e.g., cancer) or other events (e.g., pregnancy), this information may be sparse, missing, or inexact. 
For example HbA1c may be measured infrequently and or missing for some patients.
Without accurate stage information, CDSV may find invalid causes. Future work is needed to determine how such information could be obtained automatically as well as how to learn a weighting function from data. While we show that linear weighting functions perform well, it may be possible to improve upon this with domain-specific functions. Finally, while our approach performed well on real-world data, these data provide challenges for evaluation, due to the lack of true ground truth. In future work, we plan to work with clinicians to validate our causal discoveries.

\section{Conclusion}
Causal relationships vary across stages in many health applications, from pregnancy to ICU severity scores. However, prior methods for causal inference from observational time-series data cannot reliably infer such stage-dependent relationships. While some methods find spurious causal relationships, others fail to find genuine ones. To address this gap, we introduce Causal Discovery with Stage Variables (CDSV), which learns stage-dependent causal relationships while allowing information sharing across stages. This allows more true causal relationships to be recovered while avoiding spurious inferences.

We demonstrate across a variety of simulations that CDSV outperforms the state of the art (PCMCI, PCMCI+, and tsFCI). When discovering causes of low oxygen saturation in the eICU dataset, CDSV is better at  eliminating false discoveries in comparison to PCMCI and PCMCI+, which both discover incorrect inhibitory causes. In real world application to MIMIC-III to find causes of high blood pressure with the stage variable SAPS-II score, CDSV recovers more medically justified causes while PCMCI and PCMCI+ discover many causes not supported by the literature, and tsFCI fails to find any causes of BP. CDSV results in more accurate causal discovery in real-world health applications where causal relationships change through the course of a disease.

\bibliographystyle{plainnat}
\bibliography{main}

\begin{thebibliography}{38}
\providecommand{\natexlab}[1]{#1}
\providecommand{\url}[1]{\texttt{#1}}
\expandafter\ifx\csname urlstyle\endcsname\relax
  \providecommand{\doi}[1]{doi: #1}\else
  \providecommand{\doi}{doi: \begingroup \urlstyle{rm}\Url}\fi

\bibitem[Ancona et~al.(2004)Ancona, Marinazzo, and
  Stramaglia]{ancona2004radial}
Nicola Ancona, Daniele Marinazzo, and Sebastiano Stramaglia.
\newblock Radial basis function approach to nonlinear granger causality of time
  series.
\newblock \emph{Physical Review E}, 70\penalty0 (5):\penalty0 056221, 2004.

\bibitem[Bareinboim and Pearl(2013)]{bareinboim2013general}
Elias Bareinboim and Judea Pearl.
\newblock A general algorithm for deciding transportability of experimental
  results.
\newblock \emph{Journal of Causal Inference}, 1\penalty0 (1):\penalty0
  107--134, 2013.

\bibitem[Bareinboim and Pearl(2014)]{bareinboim2014transportability}
Elias Bareinboim and Judea Pearl.
\newblock Transportability from multiple environments with limited experiments:
  Completeness results.
\newblock \emph{NeurIPS}, 2014.

\bibitem[Dahhan et~al.(2009)Dahhan, Jamil, Al-Tarifi, Abouchala, and
  Kherallah]{dahhan2009validation}
T~Dahhan, M~Jamil, A~Al-Tarifi, N~Abouchala, and M~Kherallah.
\newblock Validation of the apache iv scoring system in patients with severe
  sepsis and comparison with the apache ii system.
\newblock \emph{Critical care}, 13:\penalty0 1--2, 2009.

\bibitem[Entner and Hoyer(2010)]{entner2010causal}
Doris Entner and Patrik~O Hoyer.
\newblock {On causal discovery from time series data using FCI}.
\newblock \emph{Probabilistic graphical models}, 2010.

\bibitem[Gao et~al.(2020)Gao, Xiao, Wang, Tang, Glass, and
  Sun]{10.1145/3366423.3380136}
Junyi Gao, Cao Xiao, Yasha Wang, Wen Tang, Lucas~M. Glass, and Jimeng Sun.
\newblock {StageNet: Stage-Aware Neural Networks for Health Risk Prediction}.
\newblock In \emph{Proceedings of The Web Conference 2020}, WWW '20, page
  530–540, New York, NY, USA, 2020. Association for Computing Machinery.
\newblock ISBN 9781450370233.
\newblock \doi{10.1145/3366423.3380136}.
\newblock URL \url{https://doi.org/10.1145/3366423.3380136}.

\bibitem[Gormsen et~al.(2019)Gormsen, S{\o}ndergaard, Christensen, Br{\o}sen,
  Jessen, and Nielsen]{gormsen2019metformin}
Lars~C Gormsen, Esben S{\o}ndergaard, Nana~L Christensen, Kim Br{\o}sen, Niels
  Jessen, and S{\o}ren Nielsen.
\newblock Metformin increases endogenous glucose production in non-diabetic
  individuals and individuals with recent-onset type 2 diabetes.
\newblock \emph{Diabetologia}, 62\penalty0 (7):\penalty0 1251--1256, 2019.

\bibitem[Granger(1969)]{granger1969investigating}
Clive~WJ Granger.
\newblock Investigating causal relations by econometric models and
  cross-spectral methods.
\newblock \emph{Econometrica: Journal of the Econometric Society}, pages
  424--438, 1969.

\bibitem[Greene and Sobin(2008)]{greene2008staging}
Frederick~L Greene and Leslie~H Sobin.
\newblock The staging of cancer: a retrospective and prospective appraisal.
\newblock \emph{CA: a cancer journal for clinicians}, 58\penalty0 (3):\penalty0
  180--190, 2008.

\bibitem[Grzegorczyk and Husmeier(2009)]{grzegorczyk2009non}
Marco Grzegorczyk and Dirk Husmeier.
\newblock Non-stationary continuous dynamic bayesian networks.
\newblock In \emph{NeurIPS}, 2009.

\bibitem[Hansson and Jonsson(1994)]{hansson1994logic}
Hans Hansson and Bengt Jonsson.
\newblock A logic for reasoning about time and reliability.
\newblock \emph{Formal aspects of computing}, 6\penalty0 (5):\penalty0
  512--535, 1994.

\bibitem[Huang et~al.(2020)Huang, Zhang, Zhang, Ramsey, Sanchez-Romero,
  Glymour, and Sch{\"o}lkopf]{huang2020causal}
Biwei Huang, Kun Zhang, Jiji Zhang, Joseph Ramsey, Ruben Sanchez-Romero, Clark
  Glymour, and Bernhard Sch{\"o}lkopf.
\newblock Causal discovery from heterogeneous/nonstationary data.
\newblock \emph{The Journal of Machine Learning Research}, 21\penalty0
  (1):\penalty0 3482--3534, 2020.

\bibitem[Husmeier et~al.(2010)Husmeier, Dondelinger, and
  Lebre]{husmeier2010inter}
Dirk Husmeier, Frank Dondelinger, and Sophie Lebre.
\newblock Inter-time segment information sharing for non-homogeneous dynamic
  bayesian networks.
\newblock In \emph{NeurIPS}, 2010.

\bibitem[Johnson et~al.(2016)Johnson, Pollard, Shen, Li-Wei, Feng, Ghassemi,
  Moody, Szolovits, Celi, and Mark]{johnson2016mimic}
Alistair~EW Johnson, Tom~J Pollard, Lu~Shen, H~Lehman Li-Wei, Mengling Feng,
  Mohammad Ghassemi, Benjamin Moody, Peter Szolovits, Leo~Anthony Celi, and
  Roger~G Mark.
\newblock {MIMIC-III, a freely accessible critical care database}.
\newblock \emph{Scientific data}, 3\penalty0 (1):\penalty0 1--9, 2016.

\bibitem[Johnson et~al.(2018)Johnson, Stone, Celi, and
  Pollard]{johnson2018mimic}
Alistair~EW Johnson, David~J Stone, Leo~A Celi, and Tom~J Pollard.
\newblock {The MIMIC Code Repository: enabling reproducibility in critical care
  research}.
\newblock \emph{JAMIA}, 25\penalty0 (1):\penalty0 32--39, 2018.

\bibitem[Kalisch and B{\"u}hlman(2007)]{kalisch2007estimating}
Markus Kalisch and Peter B{\"u}hlman.
\newblock {Estimating high-dimensional directed acyclic graphs with the
  PC-algorithm}.
\newblock \emph{JMLR}, 8\penalty0 (3), 2007.

\bibitem[Karam et~al.(1989)Karam, Lever, and Healy]{karam1989hypertensive}
Roger Karam, Harry~M Lever, and Bernadine~P Healy.
\newblock Hypertensive hypertrophic cardiomyopathy or hypertrophic
  cardiomyopathy with hypertension?: a study of 78 patients.
\newblock \emph{Journal of the American College of Cardiology}, 13\penalty0
  (3):\penalty0 580--584, 1989.

\bibitem[Kleinberg(2013)]{kleinberg2013causality}
Samantha Kleinberg.
\newblock \emph{Causality, Probability, and Time}.
\newblock Cambridge University Press, 2013.

\bibitem[Le~Gall et~al.(1993)Le~Gall, Lemeshow, and Saulnier]{le1993new}
Jean-Roger Le~Gall, Stanley Lemeshow, and Fabienne Saulnier.
\newblock {A new simplified acute physiology score (SAPS II) based on a
  European/North American multicenter study}.
\newblock \emph{{JAMA}}, 270\penalty0 (24):\penalty0 2957--2963, 1993.

\bibitem[Lee and Honavar(2013)]{lee2013m}
Sanghack Lee and Vasant Honavar.
\newblock m-transportability: Transportability of a causal effect from multiple
  environments.
\newblock In \emph{AAAI}, 2013.

\bibitem[Lewis et~al.(2018)Lewis, Pritchard, Evans, Butler, Alderson, Smith,
  and Roberts]{lewis2018colloids}
Sharon~R Lewis, Michael~W Pritchard, David~JW Evans, Andrew~R Butler, Phil
  Alderson, Andrew~F Smith, and Ian Roberts.
\newblock Colloids versus crystalloids for fluid resuscitation in critically
  ill people.
\newblock \emph{Cochrane Database of Systematic Reviews}, 8\penalty0 (8), 2018.

\bibitem[Marinazzo et~al.(2011)Marinazzo, Liao, Chen, and
  Stramaglia]{marinazzo2011nonlinear}
Daniele Marinazzo, Wei Liao, Huafu Chen, and Sebastiano Stramaglia.
\newblock Nonlinear connectivity by granger causality.
\newblock \emph{Neuroimage}, 58\penalty0 (2):\penalty0 330--338, 2011.

\bibitem[Murphy(2002)]{murphy2002dynamic}
Kevin~Patrick Murphy.
\newblock \emph{Dynamic bayesian networks: representation, inference and
  learning}.
\newblock PhD thesis, University of California, Berkeley, 2002.

\bibitem[Pearl(2000)]{pearl2000models}
Judea Pearl.
\newblock \emph{Causality: Models, Reasoning and Inference}.
\newblock Cambridge University Press, Cambridge, UK, 2000.

\bibitem[Pearl and Bareinboim(2014)]{pearl2014external}
Judea Pearl and Elias Bareinboim.
\newblock External validity: From do-calculus to transportability across
  populations.
\newblock \emph{Statistical Science}, 29\penalty0 (4):\penalty0 579--595, 2014.

\bibitem[Pollard et~al.(2018)Pollard, Johnson, Raffa, Celi, Mark, and
  Badawi]{pollard2018eicu}
Tom~J Pollard, Alistair~EW Johnson, Jesse~D Raffa, Leo~A Celi, Roger~G Mark,
  and Omar Badawi.
\newblock The eicu collaborative research database, a freely available
  multi-center database for critical care research.
\newblock \emph{Scientific data}, 5\penalty0 (1):\penalty0 1--13, 2018.

\bibitem[Qaseem et~al.(2013)Qaseem, Hopkins~Jr, Sweet, Starkey, and
  Shekelle]{qaseem2013screening}
Amir Qaseem, Robert~H Hopkins~Jr, Donna~E Sweet, Melissa Starkey, and Paul
  Shekelle.
\newblock Screening, monitoring, and treatment of stage 1 to 3 chronic kidney
  disease: a clinical practice guideline from the american college of
  physicians.
\newblock \emph{Annals of internal medicine}, 159\penalty0 (12):\penalty0
  835--847, 2013.

\bibitem[Robinson and Hartemink(2008)]{robinson2008non}
Joshua Robinson and Alexander Hartemink.
\newblock Non-stationary dynamic bayesian networks.
\newblock \emph{Advances in neural information processing systems}, 21, 2008.

\bibitem[Robinson et~al.(2010)Robinson, Hartemink, and
  Ghahramani]{robinson2010learning}
Joshua~W Robinson, Alexander~J Hartemink, and Zoubin Ghahramani.
\newblock {Learning Non-Stationary Dynamic Bayesian Networks}.
\newblock \emph{JMLR}, 11\penalty0 (12), 2010.

\bibitem[Rothwell et~al.(2010)Rothwell, Howard, Dolan, O'Brien, Dobson,
  Dahl{\"o}f, Poulter, Sever, et~al.]{rothwell2010effects}
Peter~M Rothwell, Sally~C Howard, Eamon Dolan, Eoin O'Brien, Joanna~E Dobson,
  Bjorn Dahl{\"o}f, Neil~R Poulter, Peter~S Sever, et~al.
\newblock Effects of $\beta$ blockers and calcium-channel blockers on
  within-individual variability in blood pressure and risk of stroke.
\newblock \emph{The Lancet Neurology}, 9\penalty0 (5):\penalty0 469--480, 2010.

\bibitem[Runge(2020)]{runge2020discovering}
Jakob Runge.
\newblock Discovering contemporaneous and lagged causal relations in
  autocorrelated nonlinear time series datasets.
\newblock In \emph{UAI}, 2020.

\bibitem[Runge et~al.(2019)Runge, Nowack, Kretschmer, Flaxman, and
  Sejdinovic]{runge2019detecting}
Jakob Runge, Peer Nowack, Marlene Kretschmer, Seth Flaxman, and Dino
  Sejdinovic.
\newblock Detecting and quantifying causal associations in large nonlinear time
  series datasets.
\newblock \emph{Science Advances}, 5\penalty0 (11):\penalty0 eaau4996, 2019.

\bibitem[Shafiee~Kamalabad and Grzegorczyk(2020)]{shafiee2020non}
Mahdi Shafiee~Kamalabad and Marco Grzegorczyk.
\newblock Non-homogeneous dynamic bayesian networks with edge-wise sequentially
  coupled parameters.
\newblock \emph{Bioinformatics}, 36\penalty0 (4):\penalty0 1198--1207, 2020.

\bibitem[Spirtes and Glymour(1991)]{spirtes1991algorithm}
Peter Spirtes and Clark Glymour.
\newblock An algorithm for fast recovery of sparse causal graphs.
\newblock \emph{Social Science Computer Review}, 9\penalty0 (1):\penalty0
  62--72, 1991.

\bibitem[Spirtes et~al.(2000)Spirtes, Glymour, and
  Scheines]{spirtes2000causation}
Peter Spirtes, Clark~N Glymour, and Richard Scheines.
\newblock \emph{Causation, prediction, and search}.
\newblock MIT press, 2000.

\bibitem[Wang et~al.(2020)Wang, McDermott, Chauhan, Ghassemi, Hughes, and
  Naumann]{wang2020mimic}
Shirly Wang, Matthew~BA McDermott, Geeticka Chauhan, Marzyeh Ghassemi,
  Michael~C Hughes, and Tristan Naumann.
\newblock Mimic-extract: A data extraction, preprocessing, and representation
  pipeline for mimic-iii.
\newblock In \emph{Proceedings of the ACM Conference on Health, Inference, and
  Learning (CHIL)}, pages 222--235, 2020.

\bibitem[Zhang et~al.(2012)Zhang, Peters, Janzing, and
  Sch{\"o}lkopf]{zhang2012kernel}
Kun Zhang, Jonas Peters, Dominik Janzing, and Bernhard Sch{\"o}lkopf.
\newblock Kernel-based conditional independence test and application in causal
  discovery.
\newblock \emph{arXiv preprint arXiv:1202.3775}, 2012.

\bibitem[Zimmerman et~al.(2006)Zimmerman, Kramer, McNair, and
  Malila]{zimmerman2006acute}
Jack~E Zimmerman, Andrew~A Kramer, Douglas~S McNair, and Fern~M Malila.
\newblock Acute physiology and chronic health evaluation (apache) iv: hospital
  mortality assessment for today’s critically ill patients.
\newblock \emph{Critical care medicine}, 34\penalty0 (5):\penalty0 1297--1310,
  2006.

\end{thebibliography}

\appendix

\section{CDSV Data Consistency Proof}\label{apd:data-consistent}

\begin{theorem}[Weighting Function Data Consistency]
    \label{thm:data-consistent}
    Given cause $c$, effect $e$, and $M$ stages with number of samples in each stage $\{N_1, \ldots, N_M\}$, the weighting function \ref{eq:sample-weight} will result in the causal significance in a stage $s: \epsilon_{avg}(c, e, s)$ to converge to its true unweighted value $\epsilon_s(c, e)$ as $\{N_1, \ldots, N_M\} \rightarrow \infty$.
    \begin{proof}
        Following the CDSV algorithm \ref{alg:cdsv}, the unweighted causal significance for stage $o$ is:
        \begin{gather*}
            \epsilon_{o}(c, e) = 
            \frac{\sum_{x \in X} P(e|c \land x) - P(e|\neg c \land x)}{|X \setminus c|} \\
        \end{gather*}
        First compute the unweighted causal significance of each stage:
        \begin{gather*}
            \epsilon_1(c, e), \epsilon_2(c, e), \ldots, \epsilon_M(c, e)
        \end{gather*}
        Now compute the CDSV average causal significance for stage $s$:
        \begin{gather*}
            \epsilon_{avg}(c, e, s) = \epsilon_1(c, e) w(s, 1) + \epsilon_2(c, e) w(s, 2) + \\
            \ldots + \epsilon_M(c, e) w(s, M) \\
            w(s, o) = 
            \begin{cases}
                \left( 1 - \frac{|s-o|}{T} \right) \cdot \frac{N_o}{N_s} & \text{if } N_s \le \tau \\
                \ \ 1 & \text{if } N_s > \tau, s = o \\
                \ \ 0 & \text{if } N_s > \tau, s \ne o
            \end{cases}
            % \\
            % \epsilon_{avg}(c, e, s) = \epsilon_1(c, e) \left(1 - \frac{|s-1|}{M}\right) \frac{N_1}{N_s} \\
            % + \epsilon_2(c, e) \left(1 - \frac{|s-2|}{M}\right) \frac{N_2}{N_s} \\
            %  \qquad \vdots \\
            % + \epsilon_M(c, e) \left(1 - \frac{|s-M|}{M}\right) \frac{N_M}{N_s}
        \end{gather*}
        Take the limit of this average causal significance as $\{N_1, \ldots, N_M\} \rightarrow \infty$:
        \begin{gather*}
            \lim_{\{N_1, \ldots, N_M\} \rightarrow \infty} \epsilon_{avg}(c, e, s) \\
            = \lim_{\{N_1, \ldots, N_M\} \rightarrow \infty} \epsilon_1(c, e) w(s, 1) + \epsilon_2(c, e) w(s, 2) \\
            + \ldots + \epsilon_M(c, e) w(s, M) \\
            = \epsilon_1(c, e) \lim_{N_1, N_s \rightarrow \infty} w(s, 1) 
            + \epsilon_2(c, e) \lim_{N_2, N_s \rightarrow \infty} w(s, 2) \\
            + \ldots + \epsilon_M(c, e) \lim_{N_M, N_s \rightarrow \infty} w(s, M)
        \end{gather*}
        Note that for any stage $s$ and different stage $o$, as $N_s \rightarrow \infty$, $w(s, o) = 0$ because $N_s > \tau$. When $s = o$, then $w(s, o) = 1$ as $N_s \rightarrow \infty$ because $N_s > \tau$. The resulting limit converges to $\epsilon_s(c, e)$:
        \begin{gather*}
            \epsilon_1(c, e) \lim_{N_1, N_s \rightarrow \infty} w(s, 1) 
            + \epsilon_2(c, e) \lim_{N_2, N_s \rightarrow \infty} w(s, 2) \\
            + \ldots + \epsilon_M(c, e) \lim_{N_M, N_s \rightarrow \infty} w(s, M) \\
            =\epsilon_1(c, e) \cdot 0 + \epsilon_2(c, e) \cdot 0 + \ldots + \epsilon_s(c, e) \cdot 1 \\
            + \ldots + \epsilon_M(c, e) \cdot 0 = \epsilon_s(c, e)
        \end{gather*}
        % Now take the limit of the previous causal significance as the number of samples in stage $s$ goes to infinity:
        % \begin{align*}
        %     \lim_{N_s \rightarrow \infty} \epsilon_{avg}(c, e, s) &= \\
        %     \lim_{N_s \rightarrow \infty} &\epsilon_1(c, e) \left(1 - \frac{|s-1|}{M}\right) \frac{N_1}{N_s} \\
        %     &+ \epsilon_2(c, e) \left(1 - \frac{|s-2|}{M}\right) \frac{N_2}{N_s} \\
        %     & \qquad \vdots \\
        %     &+ \epsilon_s(c, e) \left(1 - \frac{|s-s|}{M}\right) \frac{N_s}{N_s} \\
        %     & \qquad \vdots \\
        %     &+ \epsilon_M(c, e) \left(1 - \frac{|s-M|}{M}\right) \frac{N_M}{N_s} \\
        %     &= 0 + 0 + \ldots + \epsilon_s(c, e) + \ldots + 0 \\
        %     &= \epsilon_s(c, e)
        % \end{align*}
    \end{proof}
\end{theorem}

\section{Method Parameters} \label{apd:method-params}

\subsection{Simple Simulation}

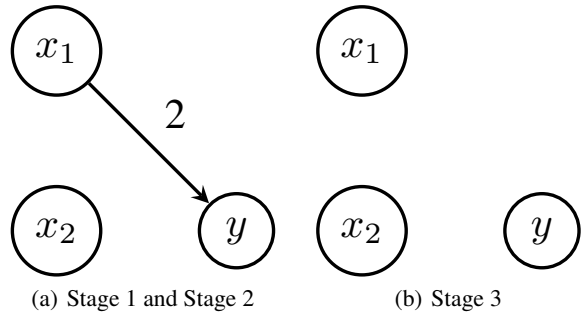
\begin{figure}[ht]
    \centering
    \subfigure[Stage 1 and Stage 2]{
        \resizebox{0.23\textwidth}{!}{
        \begin{tikzpicture}[> = stealth, auto, node distance = 1.5cm, thick]
            \tikzstyle{every state}=[
                draw = black,
                thick,
                fill = white,
                minimum size = 4mm
            ]
            \node[state] (1) {$x_1$};
            \node[state] (2) [below of=1] {$x_2$};
            \node[state] (3) [right of=2] {$y$};
            \path[->] (1) edge node {2} (3);
        \end{tikzpicture}
        }
        \label{fig:simpleSimStructStage1}
    }
    \subfigure[Stage 3]{
        \resizebox{0.23\textwidth}{!}{
        \begin{tikzpicture}[> = stealth, auto, node distance = 1.5cm, thick]
            \tikzstyle{every state}=[
                draw = black,
                thick,
                fill = white,
                minimum size = 4mm
            ]
            \node[state] (1) {$x_1$};
            \node[state] (2) [below of=1] {$x_2$};
            \node[state] (3) [right of=2] {$y$};
        \end{tikzpicture}
        }
        \label{fig:simpleSimStructStage3}
    }
    \caption{Simple Three Stage Simulation Structure}
    \label{fig:simpleSimStruct}
\end{figure}

\begin{itemize}
    \item Structure: Figure \ref{fig:simpleSimStruct}
    \item Data: 5 time-series in each stage of 1000 time points
    \item Causes = $\{x_1, x_2\}$
    \item Effects = $\{y\}$
    \item Stages = $\{1, 2, 3\}$
    \item Time Window = $(1, 4)$
    \item FDR level = $0.05$
    \item PCMCI Conditional Independence Test: Partial correlation test
\end{itemize}

\subsection{Flipped Two Stage Simulation}

\begin{figure}[ht]
    \centering
    \subfigure[Stage 1]{
        \resizebox{0.23\textwidth}{!}{
        \begin{tikzpicture}[> = stealth, auto, node distance = 1.5cm, thick]
            \tikzstyle{every state}=[
                draw = black,
                thick,
                fill = white,
                minimum size = 4mm
            ]
            \node[state] (1) {$x_1$};
            \node[state] (2) [below of=1] {$x_2$};
            \node[state] (3) [right of=2] {$y$};
            \path[->] (1) edge node {$+$} (3);
        \end{tikzpicture}
        }
        \label{fig:flippedSimStructStage1}
    }
    \subfigure[Stage 2]{
        \resizebox{0.23\textwidth}{!}{
        \begin{tikzpicture}[> = stealth, auto, node distance = 1.5cm, thick]
            \tikzstyle{every state}=[
                draw = black,
                thick,
                fill = white,
                minimum size = 4mm
            ]
            \node[state] (1) {$x_1$};
            \node[state] (2) [below of=1] {$x_2$};
            \node[state] (3) [right of=2] {$y$};
            \path[->] (1) edge node {$-$} (3);
        \end{tikzpicture}
        }
        \label{fig:flippedSimStructStage2}
    }
    \caption{Flipped Two Stage Simulation Structure}
    \label{fig:flippedSimStruct}
\end{figure}
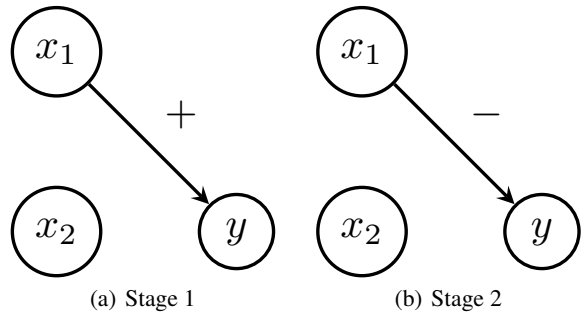

\begin{itemize}
    \item Structure: Figure \ref{fig:flippedSimStruct}
    \item Data: 5 time-series of 1000 time points in each stage
    \item Causes = $\{x_1, x_2\}$
    \item Effects = $\{y\}$
    \item Stages = $\{1, 2\}$
    \item Time Window = $(1, 4)$
    \item FDR level = $0.05$
    \item PCMCI Conditional Independence Test: Partial correlation test
\end{itemize}

\subsection{Complex Three Stage Simulation}

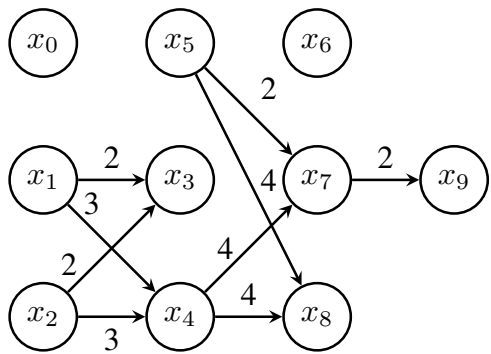
\begin{figure}[h]
    \centering
    \resizebox{0.4\textwidth}{!}{
    \begin{tikzpicture}[> = stealth, auto, node distance = 1.5cm, thick]
        \tikzstyle{every state}=[
            draw = black,
            thick,
            fill = white,
            minimum size = 4mm
        ]
        \node[state] (0) {$x_0$};
        \node[state] (1) [below of=0] {$x_1$};
        \node[state] (2) [below of=1] {$x_2$};
        \node[state] (3) [right of=1] {$x_3$};
        \node[state] (4) [right of=2] {$x_4$};
        \node[state] (5) [right of=0] {$x_5$};
        \node[state] (6) [right of=5] {$x_6$};
        \node[state] (7) [right of=3] {$x_7$};
        \node[state] (8) [right of=4] {$x_8$};
        \node[state] (9) [right of=7] {$x_9$};
        \path[->] (1) edge node {2} (3);
        \path[->] (1) edge node[above=14pt, left=0pt] {3} (4);
        \path[->] (2) edge node[below=5pt, left=7pt] {2} (3);
        \path[->] (2) edge node[below] {3} (4);
        \path[->] (4) edge node[above=0pt, left=1pt] {4} (7);
        \path[->] (4) edge node {4} (8);
        \path[->] (5) edge node {2} (7);
        \path[->] (5) edge node[right=0pt] {4} (8);
        \path[->] (7) edge node {2} (9);
    \end{tikzpicture}
    }
    \caption{Complex Three Stage Simulation Structure. This simulation has the same structure in all three stages, with the strongest relationships in stage 1, weaker in stage 2, and weakest in stage 3.}
    \label{fig:complexSimStructApd}
\end{figure}

\begin{itemize}
    \item Structure: Figure \ref{fig:complexSimStructApd}
    \item Data: 5 time-series of 1000 time points in each stage
    \item Causes = $\{x_0, x_1, x_2, x_3, x_4, x_5, x_6, x_7, x_8\}$
    \item Effects = $\{x_3, x_4, x_7, x_8, x_9\}$
    \item Stages = $\{1, 2, 3\}$
    \item Time Window = $(1, 4)$
    \item FDR level = $0.05$
    \item PCMCI Conditional Independence Test: Partial correlation test
\end{itemize}

\subsection{Complex Small Sample Simulation}

\begin{itemize}
    \item Structure: Figure \ref{fig:complexSimStructApd}
    \item Data: 1 time-series in stage 1 with 1000 time points, 5 time-series in stage 2 with 1000 time points, 5 time-series in stage 3 with 1000 time points
    \item Causes = $\{x_0, x_1, x_2, x_3, x_4, x_5, x_6, x_7, x_8\}$
    \item Effects = $\{x_3, x_4, x_7, x_8, x_9\}$
    \item Stages = $\{1, 2, 3\}$
    \item Time Window = $(1, 4)$
    \item FDR level = $0.05$
    \item PCMCI Conditional Independence Test: Partial correlation test
\end{itemize}

\section{MIMIC High Blood Pressure Experiment Details} \label{apd:mimic-experiment}

\subsection{Causes, Effects and Parameters}

\begin{itemize}
    \item Causes = $\{$ventilator, vasopressor, adenosine, dobutamine,
                   dopamine, epinephrine, isuprel, milrinone, norepinephrine, phenylephrine, vasopressin, colloid bolus, crystalloid bolus, non-invasive ventilation, high heart rate, normal heart rate, low heart rate, high respiratory rate, normal respiratory rate, low respiratory rate, normal oxygen saturation, low oxygen saturation$\}$
    \item Effects = high blood pressure
    \item Stages = High SAPS-II score, Medium SAPS-II score, Low SAPS-II score
    \item Time Window = $(1, 4)$ hours
    \item FDR level = $0.05$
    \item PCMCI Conditional Independence Test: Partial correlation test
\end{itemize}

\subsection{High Blood Pressure Experiment Variable Categories}

To categorize each variable in this experiment, we use empirical quantiles. We define the $0-20\%$ data as low, $20-80\%$ as normal/medium, and $80-100\%$ as high. Below we have provided the ranges of each variable of interest: 
\begin{itemize}
    \item Mean Blood Pressure: $0-60$, $60-100$, $>100$
    \item SAPS-II Score: $0-22$, $22-48$, $48-163$
    \item Heart Rate: $0-60$, $60-110$, $>110$
    \item Respiratory Rate: $0-12$, $12-21$, $>21$
\end{itemize}

\section{eICU Low Oxygen Saturation Experiment Details} \label{apd:eicu-experiment}

\subsection{Causes, Effects and Parameters}

\begin{itemize}
    \item Causes = $\{$vasopressor, oxygenization,
                    high central venous pressure, normal central venous pressure, low central venous pressure, high heart rate, normal heart rate, low heart rate, high respiratory rate, normal respiratory rate, low respiratory rate, normal oxygen saturation, high temperature, normal temperature, low temperature, high blood pressure, normal blood pressure, low blood pressure$\}$
    \item Effects = low oxygen saturation
    \item Stages = High APACHE IV score, Low APACHE IV score
    \item Time Window = $(1, 4)$ hours
    \item FDR level = $0.05$
    \item PCMCI Conditional Independence Test: Partial correlation test
\end{itemize}

\subsection{Low Oxygen Saturation Experiment Variable Categories}

To categorize each variable in this experiment, we use empirical quantiles. We define the $0-20\%$ data as low, $20-80\%$ as normal/medium, and $80-100\%$ as high. Below we have provided the ranges of each variable of interest: 
\begin{itemize}
    \item APACHE IV Score: $0-70$, $\ge 70$
    \item Central Venous Pressure: $0-6$, $6-20$, $> 20$
    \item Heart Rate: $0-70$, $70-101$, $>101$
    \item Respiratory Rate: $0-15$, $15-25$, $>25$
    \item Oxygen Saturation: $0-94.5$, $\ge 94.5$
    \item Temperature: $0-36.6$, $36.6-37.9$, $\ge 37.9$
    \item Total Blood Pressure: $0-163$, $163-211$, $>211$
\end{itemize}

\section{Related Work Comparison} \label{apd:related-work}

\begin{table*}[h]
\begin{tabular}{@{}lccccc@{}}
\toprule
 & \multicolumn{1}{p{3cm}}{\centering Stage Dependent \\ Causal Relationships}
 & \multicolumn{1}{p{3cm}}{\centering Transfer Between \\ Stages} 
 & \multicolumn{1}{p{2cm}}{\centering Exact \\ Inference} 
 & \multicolumn{1}{p{3cm}}{\centering Prior Probability \\ Not Required} 
 & \multicolumn{1}{p{2cm}}{\centering Causal Graph \\ Not Required} \\ \midrule
SCM & \xmark & \xmark & \cmark & \xmark & \xmark \\
DBN & \xmark & \xmark & \xmark & \xmark & \xmark \\
nsDBN & \cmark & \cmark & \xmark & \xmark & \xmark \\
PCMCI & \xmark & \xmark & \cmark & \cmark & \cmark \\
PCMCI+ & \xmark & \xmark & \cmark & \cmark & \cmark \\
tsFCI & \xmark & \xmark & \cmark & \cmark & \cmark \\
Kleinberg 2013 & \xmark & \xmark & \cmark & \cmark & \cmark \\
CDSV & \cmark & \cmark & \cmark & \cmark & \cmark \\ \bottomrule
\end{tabular}
\caption{A comparison of related methods and their features. CDSV is the only method that fits our desiderata.}
\label{tab:related}
\end{table*}

In Table \ref{tab:related} we compare CDSV to related methods. Note that only CDSV has support for stage-dependent causal relationships as well as our other requirements.

\section{Weighting Function Design}
\label{apd:weighting-func}

When determining the weighting function, we recommend the following guidelines:
\begin{enumerate}
    \item Review literature about the health condition/disease you are studying, looking for medically validated measures of its state.
    \item Based on the method by which the above measure is computed, determine the dynamics of the condition/disease.
    \item Once the dynamics are of the disease and corresponding metric are understood, then choose the appropriate weighting function.
\end{enumerate}

The above process is often straightforward in healthcare because many health metrics such as week of gestation, hemoglobin A1C, cancer stage, and others have been extensively studied and validated.

As an example of this, in our real data experiments we were studying adverse events in the ICU such as high blood pressure. This led us to find the SAPS II score and APACHE IV score, which were both clinically validated ICU severity scores. Since these scores have been used to assess patient severity, risk of mortality and have been calibrated \citep{dahhan2009validation}, we categorized patients into score groups and relate them with a linear weighting function.

\end{document}